\documentclass[letterpaper]{article} 
\usepackage{aaai19}  
\usepackage{times}  
\usepackage{helvet}  
\usepackage{courier}  
\usepackage{url}  
\usepackage{graphicx}  
\usepackage{subfigure}
\usepackage{amsmath}
\usepackage{booktabs}
\usepackage{amssymb}
\usepackage{color}
\usepackage{threeparttable}
\usepackage{amsfonts}
\usepackage{amsmath}
\usepackage{algorithm}
\usepackage{algorithmic}
\usepackage{multirow}
\usepackage{graphicx}
\usepackage{booktabs}       
\usepackage{nicefrac}       
\usepackage{microtype}      
\usepackage{float,makecell,booktabs,color,subfigure,caption,setspace}
\usepackage{amssymb}
\usepackage{subfig}
\usepackage{threeparttable}

\title{Diversity-Driven Extensible Hierarchical Reinforcement Learning}
\author{Yuhang~Song\textsuperscript{1}, Jianyi~Wang\textsuperscript{2,*}, Thomas~Lukasiewicz\textsuperscript{1}, Zhenghua~Xu\textsuperscript{1,3,\dag}, Mai~Xu\textsuperscript{2}\\ \textsuperscript{1} Department of Computer Science, University of Oxford, UK\\ \textsuperscript{2} School of Electronic and Information Engineering, Beihang University, China\\ \textsuperscript{3} State Key Laboratory of Reliability and Intelligence of Electrical Equipment, Hebei University of Technology, China\\ \textsuperscript{\dag} Corresponding author, email: xuzhenghua1987@gmail.com \textsuperscript{*}Co-first author\\}

\begin{document}
\maketitle

\begin{abstract}
Hierarchical reinforcement learning (HRL) has recently shown promising advances on speeding up learning, improving the exploration, and discovering intertask transferable skills. Most recent works focus on HRL with two levels, i.e., a master policy manipulates subpolicies, which in turn manipulate primitive actions. However, HRL with multiple levels is usually needed in many real-world scenarios, whose ultimate goals are highly abstract, while their actions are very primitive. Therefore, in this paper, we propose a diversity-driven extensible HRL (DEHRL), where an extensible and scalable framework is built and learned levelwise to realize HRL with multiple levels. DEHRL follows a popular assumption: diverse subpolicies are useful, i.e., subpolicies are believed to be more useful if they are more diverse. However, existing implementations of this diversity assumption usually have their own drawbacks, which makes them inapplicable to HRL with multiple levels. Consequently, we further propose a novel diversity-driven solution to achieve this assumption in DEHRL. Experimental studies evaluate DEHRL with five baselines from four perspectives in two domains; the results show that DEHRL outperforms the state-of-the-art baselines in all four aspects.
\end{abstract}

\section{Introduction}

Hierarchical reinforcement learning (HRL) recombines sequences of basic \textrm{actions} to form subpolicies \cite{sutton1999between,parr1998reinforcement,dietterich2000hierarchical}.
It can be used to speed up the learning \cite{bacon2017option}, improve the exploration to solve tasks with sparse \textit{extrinsic rewards} (i.e., rewards generated by the environment) \cite{csimcsek2004using}, or learn meta-skills that can be transferred to new problems \cite{frans2017meta}.
Although most previous approaches to HRL require hand-crafted subgoals to pretrain subpolicies \cite{heess2016learning} or extrinsic rewards as supervisory signals \cite{vezhnevets2016strategic}, the recent ones seek to discover subpolicies without manual subgoals or pretraining.
Most of them are working in a top-down fashion, such as \cite{xu2017neural,bacon2017option}, where a given \textrm{agent} first explores until it accomplishes a trajectory that reaches a positive extrinsic reward. Then, it tries to recombine the basic actions in the trajectory to build useful or reasonable~subpolicies.

However, such top-down solutions are not practical in some situations, where the extrinsic rewards are sparse, and the action space is large (called \textit{sparse extrinsic reward problems}); this is because positive extrinsic rewards are almost impossible to be reached by exploration using basic actions in such scenarios.
Therefore, more recent works focus on discovering ``useful'' subpolicies in a bottom-up fashion \cite{lakshminarayanan2016option,kompella2017continual}, which are capable of discovering subpolicies before reaching a positive extrinsic reward.
In addition, the bottom-up strategy can discover subpolicies that better facilitate learning for an unseen problem \cite{frans2017meta}.
This 
is also called 
meta-learning \cite{finn2017model}, or more precisely meta-reinforcement-learning \cite{al2017continuous}, where subpolicies shared~across different tasks are called meta-subpolicies (or meta-skills).

\begin{figure*}
	\centering
		\centerline{\includegraphics[width=1.1\textwidth]{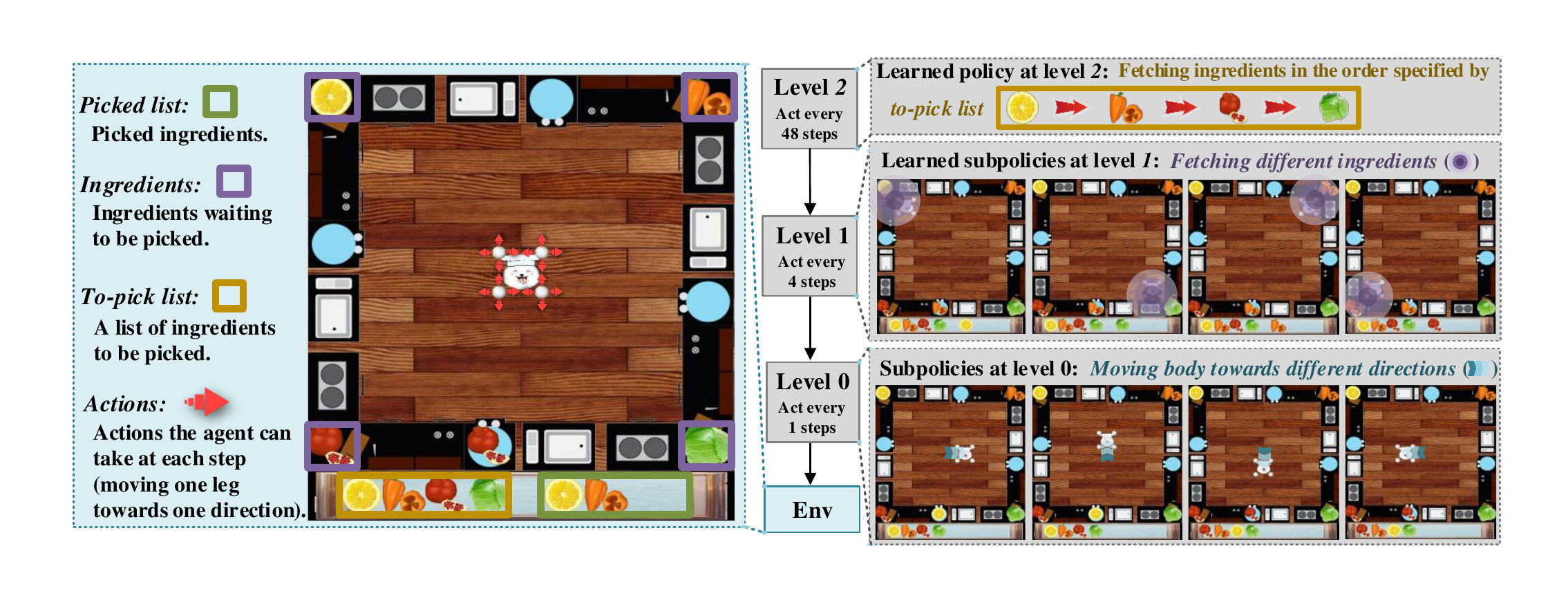}}
		\vspace{-1.5em}
		\caption{Playing OverCooked with HRL of three levels.}
		\label{overcooked}
\end{figure*}

However, none of the above methods shows the capability to build extensible HRL with multiple levels, i.e., building subpolicies upon subpolicies, which is usually needed in many real-world scenarios, whose \textit{ultimate goals} are highly abstract, while their basic \textit{actions} are very primitive.
We take the game  \textit{OverCooked} (shown in Fig.~\ref{overcooked}) as an example. The ultimate goal of OverCooked is to let an \textit{agent} fetch multiple \textit{ingredients} (green box) in a particular sequence according to a \textit{to-pick list} (yellow box), which is shuffled in every episode. However, the basic action of the agent is so primitive (thus called \textit{primitive action} in the following) that it can just move one of its four legs towards one of the four directions at each step (marked by red arrows), and the agent's body can be moved only when all four legs are moved to the same direction. Consequently, although we can simplify the task by introducing subpolicies to learn to move the body towards different directions with four steps of \textrm{primitive actions}, the ultimate goal is still difficult to be reached, because the \textrm{to-pick list} changes every episode.

Fortunately, this problem can be easily overcome if HRL has multiple levels: by defining the previous subpolicies as subpolicies at level $0$, HRL with multiple levels can build subpolicies at level $1$ to learn to fetch different \textrm{ingredients} based on subpolicies at level $0$; obviously, a policy based on subpolicies at level $1$ is capable to reach the ultimate goal more easily than that based on subpolicies at level $0$.


Motivated by the above observation, in this work, we propose a diversity-driven extensible HRL (\textit{DEHRL}) approach, which is constructed and trained levelwise (i.e., each level shares the same structure and is trained with exactly the same algorithm), making the HRL framework extensible to build higher levels. DEHRL follows a popular \textit{diversity assumption}: diverse subpolicies are useful, i.e., subpolicies are believed to be more useful if they are more diverse. Therefore, the objective of DEHRL at each level is to learn corresponding subpolicies that are as diverse as possible, thus called diversity-driven.

However, existing implementations of this diversity assumption usually have their own drawbacks, which make them inapplicable to HRL with multiple levels. For example, (i) the implementation in \cite{daniel2012hierarchical} works in a top-down fashion; (ii) the one in \cite{haarnoja2018latent} cannot operate different layers at different temporal scales to solve temporally delayed reward tasks; and (iii) the implementation in \cite{gregor2016variational,florensa2017stochastic} is not extensible to higher levels.


Consequently, we further propose a novel diversity-driven solution to achieve this assumption in DEHRL: We first introduces a \textit{predictor} at each level to dynamically predict the resulting \textrm{state} of each subpolicy. Then, the diversity assumption is achieved by giving higher \textit{intrinsic rewards} to subpolicies that result in more diverse states; consequently, subpolicies in DEHRL converge to taking actions that result in most diverse states.
Here, \textit{intrinsic rewards} are the rewards generated by the agent.



We summarize the contributions of this paper as follows:
\begin{itemize}
	\item We propose a diversity-driven extensible hierarchical reinforcement learning (DEHRL) approach. To our knowledge, DEHRL is the first learning algorithm that is built and learned levelwise with verified scalability, so that HRL with multiple levels can be realized end-to-end without human-designed extrinsic rewards.
	\item We further propose a new diversity-driven solution to implement and achieve the widely adopted diversity assumption in HRL with multiple levels. 
\item	Experimental studies evaluate DEHRL with five baselines from four perspectives in two domains. The results show that, comparing to the baselines, DEHRL achieves the following advantages: (i) DEHRL can discover useful subpolicies more effectively, (ii) DEHRL can solve the sparse extrinsic reward problem more efficiently, (iii) DEHRL can learn better intertask-transferable meta subpolicies, and (iv)~DEHRL has a good portability.
\end{itemize}

\section{Diversity-Driven Extensible HRL}


This section introduces the new DEHRL framework as well as the integrated diversity-driven solution.
The structure of DEHRL is shown in Fig.~\ref{m_structure}, where each level $l$ contains a \textit{policy} (denoted $\pi^l$), a \textit{predictor} (denoted  $\beta^l$), and an \textit{estimator}.
The above \textit{policy} and \textit{predictor} are two deep neural networks (i.e., parameterized functions) with $\pi^l$ and $\beta^l$ denoting their trainable parameters, while \textit{estimator} only contains some operations without trainable parameters.
For any two neighboring levels (e.g., level $l$ and level $l-1$), three connections are shown in Fig.~\ref{m_structure}:
\begin{itemize}
	\item The policy at the upper level $\pi^l$ produces the action $a^l_t$, which is treated as an input for the policy at the lower level $\pi^{l-1}$;
	\item The predictor at the upper level $\beta^l$ makes several predictions, which are passed to the estimator at the lower level $l-1$;
	\item Using the predictions from the upper level $l$, the estimator at the lower level $l-1$ generates an intrinsic reward $b^{l-1}_t$ to train the policy at the lower level $\pi^{l-1}$;
\end{itemize}

\subsection{Policy}
\label{policy}

\begin{figure}
	\centering
		\centerline{\includegraphics[width=\columnwidth]{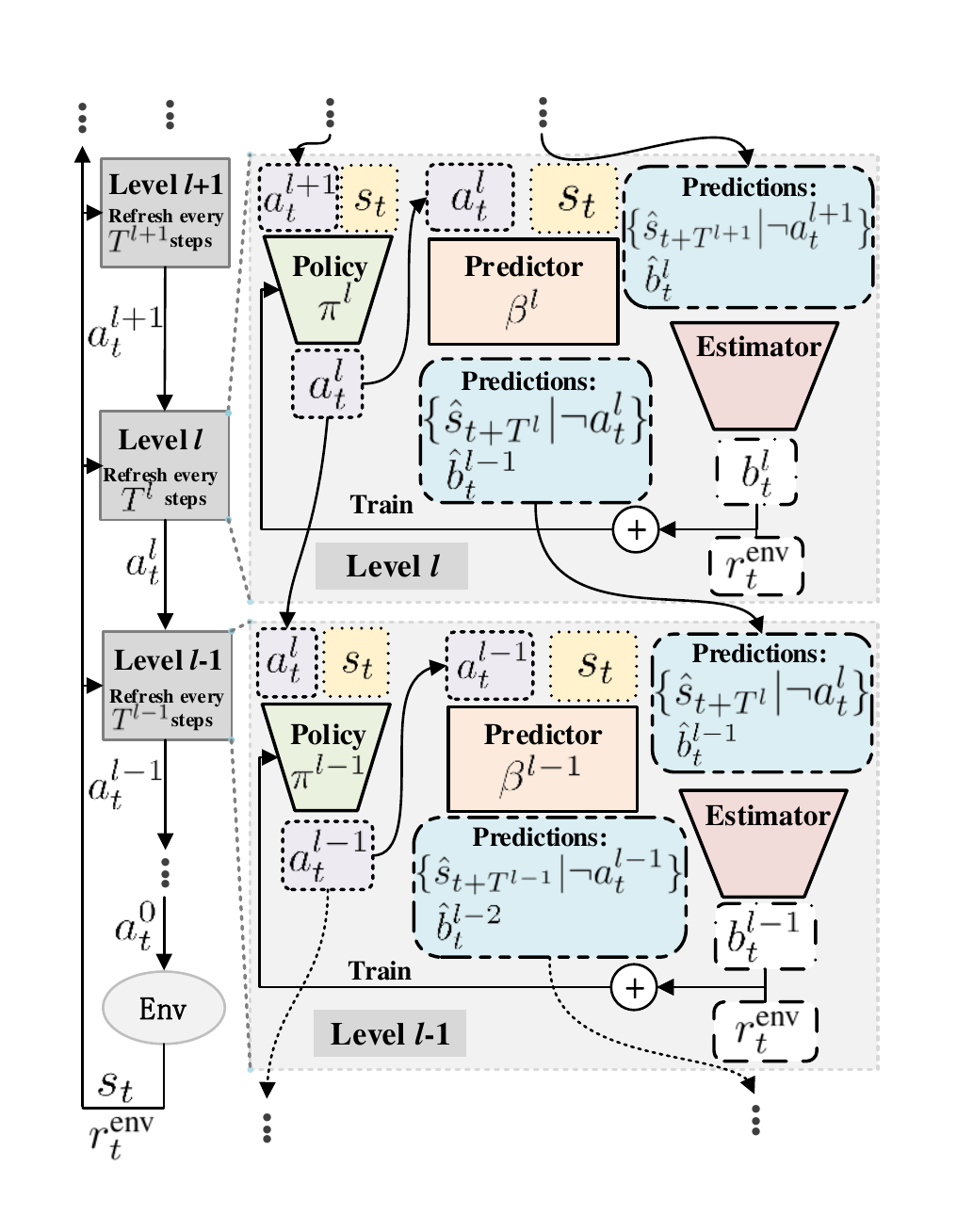}}
		\vspace{-1.5em}
		\caption{The framework of DEHRL.}
		\label{m_structure}
\end{figure}

As shown in Fig.~\ref{m_structure}, the \textrm{policies} for different levels act at different frequencies, i.e., the policy $\pi^l$ samples an action every $T^l$ steps.
Note that $T^l$ is always an integer multiple of $T^{l-1}$, and $T^0$ always equals to $1$, so the time complexity of the proposed framework does not grow linearly as the level goes higher.
$T^{l}$ for $l>0$ are hyper-parameters.
At  level~$l$, the policy $\pi^l$ takes as input the current state $s_t$ and the action from the upper level $a_t^{l+1}$, so that the output of $\pi^l$ is conditional to $a_t^{l+1}$.
Note that $a_t^{l+1} \in \mathbb{A}^{l+1}$, where $\mathbb{A}^{l+1}$ is the output action space of $\pi^{l+1}$.
Thus, $\mathbb{A}^{0}$ should be set to the action space of the \textit{environment} for the policy $\pi^0$ directly taking actions on the environment, while $\mathbb{A}^{l}$ of $l>0$ are hyper-parameters.
The policy takes as input both $s_t$ and $a_t^{l+1}$ to integrate multiple subpolicies into one model; a similar idea is presented in \cite{florensa2017stochastic}.
The detailed network structure of the policy $\pi^l$ is presented in the arXiv release\footnote{https://arxiv.org/abs/1811.04324}.
Then, the policy $\pi^l$ produces the action $a_t^l\in\mathbb{A}^{l}$ by sampling from a parameterized categorical distribution:
\begin{equation}
a_t^l = \pi^l (s_t, a_t^{l+1}).
\end{equation}

The reward to train $\pi^l$ combines the extrinsic reward from the environment $r_t^{\textrm{env}}$ and the intrinsic reward $b_t^l$ generated from the estimator at level $l$.
When facing games with very sparse extrinsic rewards, where $r_t^{\textrm{env}}$ is absent most of the time, $b_t^l$ will guide the policy at this level $\pi^l$ to learn diverse subpolicies, so that the upper level policy $\pi^{l+1}$ may reach the sparse positive extrinsic reward more easily.
The policy $\pi^l$ is trained with the PPO algorithm \cite{schulman2017proximal}, but our framework does not restrict the policy training algorithm to use.
The following denotes the loss of training the policy~$\pi^l$:
\begin{equation}
\label{policy-loss}
L^l_{\textrm{policy}} = \textrm{PPO} \left(a_t^l, (r_t^{\textrm{env}}+\lambda b_t^l) | \pi^l \right),
\end{equation}
where $\pi^l$ means that the gradients of this loss are only passed to the parameters in $\pi^l$, and $\lambda$ is a hyper-parameter set to $1$ all the time (the section on the estimator below will introduce a normalization of the intrinsic reward $b_t^l$, making $\lambda$ free from careful tuning).

\subsection{Predictor}

As shown in Fig.~\ref{m_structure}, the predictor at level $l$ (i.e., $\beta^l$) takes as input the current state $s_t$ and the action taken by the policy at level $l$ (i.e., $a_t^l$) as a one-hot vector.
The integration of $s_t$ and $a_t^l$ is accomplished by approximated multiplicative interaction \cite{oh2015action}, so that any predictions made by the predictor $\beta^l$ is conditioned on the action input of $a_t^l$.
The predictor makes two predictions, denoted $\hat{s}_{t+T^l}$ and $\hat{b}^{l-1}_t$, respectively.
Thus, the forward function of  $\beta^l$ is:
\begin{equation}
\label{predictor}
\hat{s}_{t+T^l}, \hat{b}^{l-1}_t = \beta^l (s_t, a_t^l).
\end{equation}
The detailed network structure of the predictor $\beta^l$ is given in the arXiv release.
The first prediction $\hat{s}_{t+T^l}$ in~\eqref{predictor} is trained to predict the state after $T^l$ steps with following loss function:
\begin{equation}
\label{transition-loss}
L^l_{\textrm{transition}} = \textrm{MSE} (s_{t+T^l}, \hat{s}_{t+T^l} | \beta^l),
\end{equation}
where \textrm{MSE} is the mean square error, and $\beta^l$ indicates that the gradients of this loss are only passed to the parameters in $\beta^l$.
The second prediction $\hat{b}^{l-1}_t$ in~\eqref{predictor} is trained to approximate the intrinsic reward at the lower level $b^{l-1}_t$, with the loss function
\begin{equation}
\label{intrinsic-reward-loss}
L^l_{\textrm{intrinsic reward}} = \textrm{MSE} (b^{l-1}_t, \hat{b}^{l-1}_t | \beta^l),
\end{equation}
where  $\beta^l$ means that the gradients of this loss are only passed to the parameters in $\beta^l$.
The next section about the estimator will show that the intrinsic reward $b^{l-1}_t$ is also related to the action $a^{l-1}_t$ sampled according to the policy at the lower level $\pi^{l-1}$.
Since $a^{l-1}_t$ is not fed into the predictor $\beta^l$, the intrinsic reward $\hat{b}^{l-1}_t$ is actually an estimation of the expectation of $b^{l-1}_t$ under the current $\pi^{l-1}$:
\begin{equation}
\hat{b}^{l-1}_t = \mathbb{E}_{\pi^{l-1}} \{b^{l-1}_t\}.
\end{equation}
The above two predictions will be used in the estimator, described in the following section.

The predictor is active at the same frequency as that of the policy.
Each time the predictor $\beta^l$ is active, it produces several predictions feeding to the estimator at level $l-1$, including $\hat{b}^{l-1}_t$ and $\{\hat{s}_{t+T^l} | \neg a_t^l\}$, where $\{\hat{s}_{t+T^l} | \neg a_t^l\}$ denotes a set of predictions when feeding the predictor $\beta^l$ with actions other than~$a_t^l$.

\subsection{Estimator}

As shown in Fig.~\ref{m_structure}, taking as input $\hat{b}^{l-1}_t$ and $\{\hat{s}_{t+T^l} | \neg a_t^l\}$, the estimator produces the intrinsic reward $b_t^{l-1}$, which is used to train the policy $\pi^{l-1}$, as described in the policy section.
The design of the estimator is motivated as follows:
\begin{itemize}
	\item If the currently selected subpolicy $\pi^{l-1} (s_t, a_t^{l})$ for the upper-level action $a_t^{l}$ differs from the subpolicies for other actions (i.e., $\{\pi^{l-1} (s_t, \neg a_t^{l})\}$), then the intrinsic reward $b_t^{l-1}$ should be high;
	\item The above difference can be measured via the distance between the states resulting from the subpolicy $\pi^{l-1} (s_t, a_t^{l})$ and subpolicies $\{\pi^{l-1} (s_t, \neg a_t^{l})\}$. Note that since $a_t^{l}$ is selected every $T^l$ steps, these resulting states are the ones at~$t+T^l$;
\end{itemize}
In the above motivation, the resulting state of the subpolicy $\pi^{l-1} (s_t, a_t^{l})$ is the real state environment returned after $T^l$ steps (i.e., $s_{t+T^l}$),
while the resulting states of the subpolicies $\{\pi^{l-1} (s_t, \neg a_t^{l})\}$ have been predicted by the predictor at the upper level $l$ (i.e., $\{\hat{s}_{t+T^l} | \neg a_t^l\}$), as described in the last section.
Thus, the intrinsic reward $b_t^{l-1}$ is computed as follows:
\begin{equation}
\label{intrinsic-reward}
b_t^{l-1} = \sum_{s \in \{\hat{s}_{t+T^l} | \neg a_t^l\}} D(s_{t+T^l},s),
\end{equation}
where $D$ is the distance chosen to measure the distance between states.
In practice, we combine the L1 distance and the distance between the center of mass of the states, to obtain information on color changes as well as objects moving.
A more advanced way to measure the above distance is to match features in states and to measure the movements of the matched features, or to integrate the inverse model in \cite{pathak2017curiosity} to capture the action-related feature changes.
However, the above advanced ways are not investigated here, due to the scope of the paper.
Equation \eqref{intrinsic-reward} gives a high intrinsic reward, if $s_{t+T^l}$ is far from $\{\hat{s}_{t+T^l} | \neg a_t^l\}$ overall.
In practice, we find that punishing $s_{t+T^l}$ from being too close to the one state in $\{\hat{s}_{t+T^l} | \neg a_t^l\}$ that is closest to $s_{t+T^l}$ is a much better choice.
Thus, we replace the sum in \eqref{intrinsic-reward} with the minimum, and find that it consistently gives the best intrinsic reward estimation.

Estimating the intrinsic reward with distances of high dimensional states comes with the problem that the changes in distance that we want the intrinsic reward to capture is extremely small, compared to the mean of the distances.
Thus, we use the estimation of the expectation of the intrinsic reward $b^{l-1}_t$ (i.e., $\hat{b}^{l-1}_t$ described in last section) to normalize~$b^{l-1}_t$:
\begin{equation}
b^{l-1}_t \leftarrow b^{l-1}_t - \hat{b}^{l-1}_t.
\end{equation}
In practice, this normalization gives a stable algorithm without need to tune $\lambda$ according to the distance that we choose or the convergence status of the predictor at the upper level.
We jointly optimize the loss functions of \eqref{policy-loss}, \eqref{transition-loss}, and \eqref{intrinsic-reward-loss}.

\section{Experiments}

\begin{table}
	\centering
	\caption{The settings of DEHRL.}
	
\smallskip 	
	\resizebox{0.6\columnwidth}{!}{%
		\begin{tabular}{cccccc}
			\midrule
			$\mathbb{A}^0$ & $\mathbb{A}^1$ & $\mathbb{A}^2$ & $T^0$ & $T^1$ & $T^2$  \\
			16  & 5    & 5 & 1   & 1*4  & 1*4*12 \\ \hline
		\end{tabular}%
	}
	\label{dehrl-param-table}
\end{table}

We conduct experiments to evaluate DEHRL and six baselines based on two games, OverCooked (shown in Fig.~\ref{overcooked}) and MineCraft.
The important hyper-parameters of DEHRL are summarized in Table~\ref{dehrl-param-table}, while other details (e.g., neural network architectures and hyper-parameters in the policy training algorithm) are provided in the arXiv release.
Easy-to-run codes have been released to further clarify the details and facilitate future research\footnote{https://github.com/YuhangSong/DEHRL}.
An evaluation on more domains (such as MontezumaRevenge, etc.) can also be found in this repository.

\begin{figure}
	\centering
		\centerline{\includegraphics[width=\columnwidth]{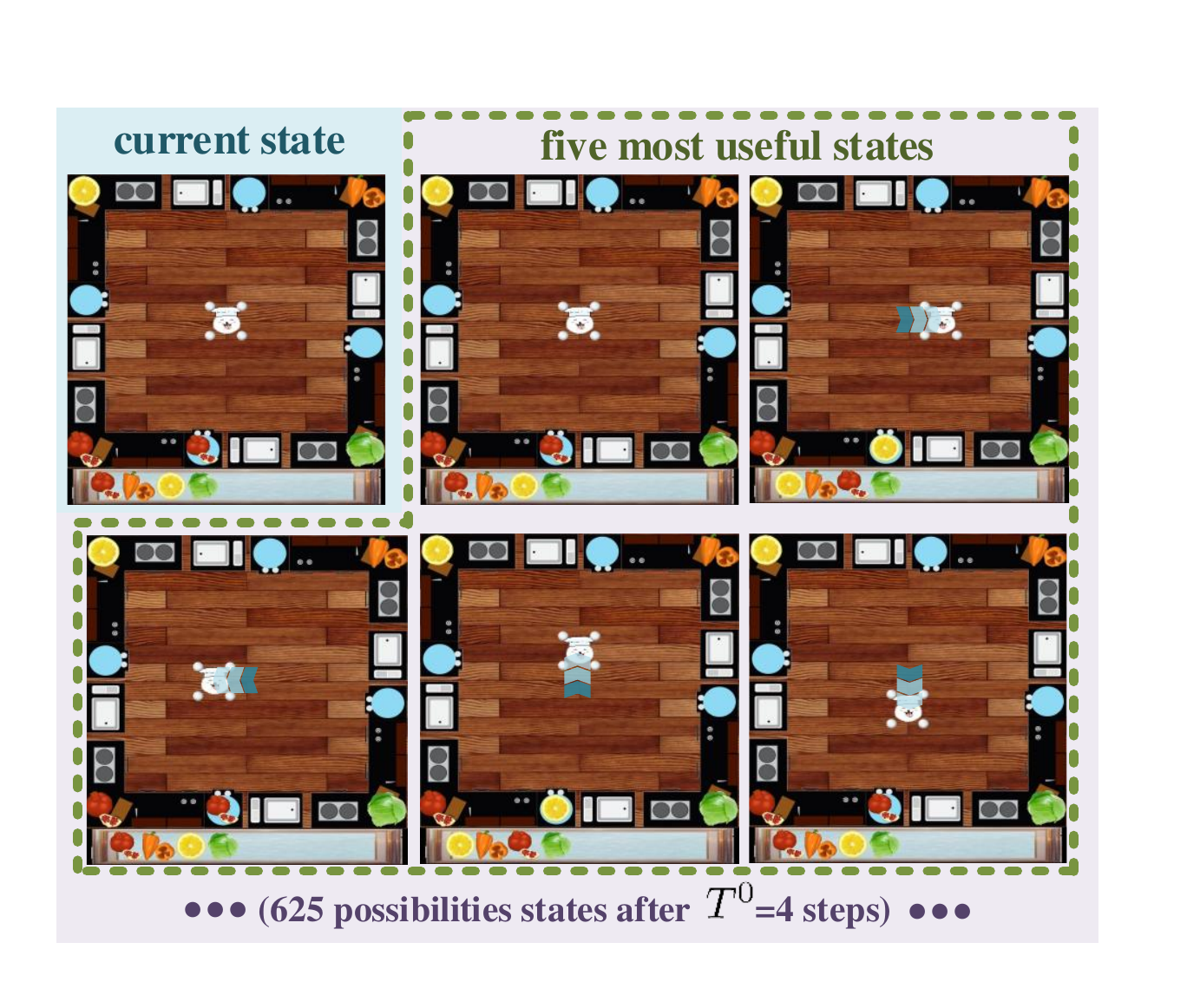}}
		\vspace{-1em}
		\caption{State examples of Overcooked.}
		\label{figure_exp}
\end{figure}

\subsection{Subpolicy Discovery}

We first evaluate DEHRL in OverCooked to see if it discovers diverse subpolicies more efficiently, compared to the state-of-the-art baselines towards option discovery \cite{florensa2017stochastic,bacon2017option}.

As shown in Fig.~\ref{overcooked}, an \textrm{agent} in OverCooked can move one of the four legs towards one of the four directions at each step, so its action space is $16$.	Only after all four legs are moved towards the same direction, the body of an \textrm{agent} can be moved towards this direction, and all four legs are then reset.
There are four different ingredients at the corners of the kitchen (marked by green box). An ingredient is automatically picked up when the agent reaches it. The left lower corner shows a list of ingredients that the chief needs to pick in sequence to complete a dish (marked by a red box), called to-pick list.

\begin{table*}
	\centering
	\caption{\textrm{The different settings of extrinsic rewards in OverCooked.}}
	
	\smallskip 
	\begin{threeparttable}
		\resizebox{\textwidth}{!}{%
			\begin{tabular}{cccc}
				\midrule
				\textit{reward-level} & \textit{goal-type: any} (easy) & \textit{goal-type: fix} (middle) & \textit{goal-type: random} (hard)  \\ \midrule
				$1$ (easy) & Get any ingredient.  & Get a particular ingredient. & \begin{tabular}{@{}c@{}}Get the first ingredient shown \\ in the shuffling* to-pick list.\end{tabular} \\
				$2$ (hard) & Get 4 ingredients in any order.  & Get 4 ingredients in a particular order. & \begin{tabular}{@{}c@{}}Get 4 ingredients in order according \\  to the shuffling* to-pick list.\end{tabular} \\ \hline
			\end{tabular}}%
			\begin{tablenotes}
				\footnotesize{\item[\hspace{10em}*] The to-pick list is shuffled every episode.}
			\end{tablenotes}
	\end{threeparttable}
	\label{game-settings}
\end{table*}


\subsubsection{Without Extrinsic Rewards.}

We first aim to discover useful subpolicies without extrinsic rewards. Since there are four legs, and every leg of an agent has five possible states (staying or moving towards four directions), there are totally $5^4=625$ possible states for every $T^0=4$ steps. As shown in Fig.~\ref{figure_exp}, five of them are the most useful states (i.e., the ones that are most diverse to each other), whose four legs have the same state, making the body of the chief move towards four directions or stay still.


Consequently, a good implementation of the diversity assumption should be able to learn subpolicies at level $0$ that can result in the five most useful states (called \emph{five useful subpolicies}) efficiently and comprehensively. Therefore, given $\mathbb{A}^1=5$ (i.e., discovering only five subpolicies), and the number of steps being $10$ millions, the five subpolicies learned by DEHRL at level $0$ are exactly the five useful subpolicies. Furthermore, \emph{SNN} \cite{florensa2017stochastic} is a state-of-the-art implementation of the diversity assumption, which is thus tested as a baseline under the same setting. However, only one of the five useful subpolicies is discovered by SNN. We then repeat experiments three times with different training seeds, and the results are the same. Furthermore, we loose the restriction by setting $\mathbb{A}^1=20$ (i.e., discovering $20$ subpolicies) and the number of steps is $20$ millions. With no surprise, the five useful subpolicies are always included in the $20$ discovered subpolicies of DEHRL; however, the $20$ subpolicies discovered by SNN still contain only one useful subpolicy.

The superior performance of DEHRL comes from the diversity-driven solution, which gives higher intrinsic rewards to subpolicies that result in more diverse states; consequently, subpolicies in DEHRL converge to taking actions that result in most diverse states. And the failure of SNN may be because the objective of SNN is to maximize mutual information, so it only guarantees to discover subpolicies resulting in different states, but these different states are not guarantees to be most diverse to each other. Similar failures are found in other state-of-the-art solutions (e.g., \cite{gregor2016variational}); we thus omit the analysis due to space limit.

\begin{figure}
	\centering
		\centerline{\includegraphics[width=\columnwidth]{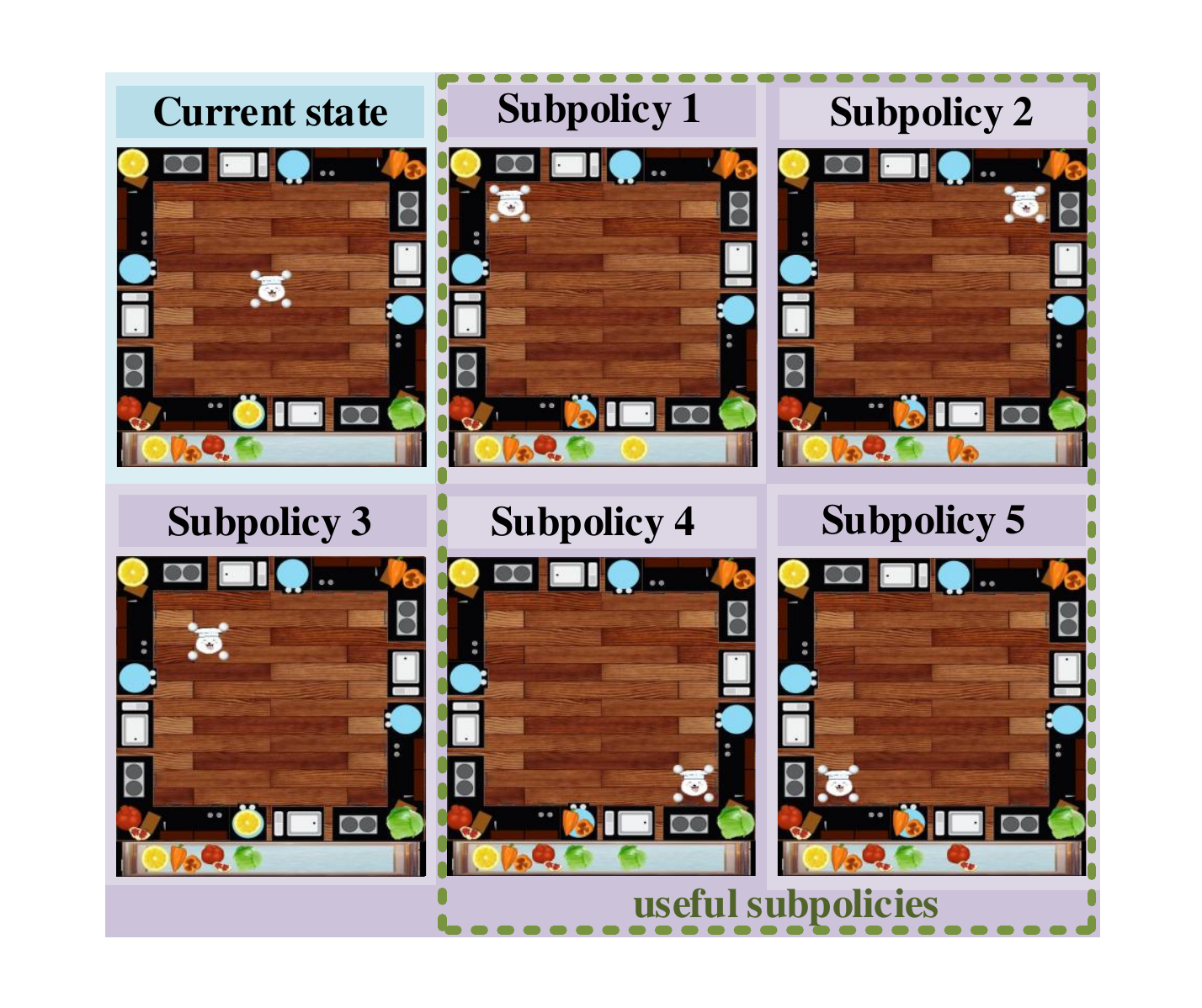}}
		\vspace{-1em}
		\caption{Subpolicies learned 
			at level 1 in DEHRL.}
		\label{sub-polices-level-1}
\end{figure}

As for finding useful subpolicies at higher levels, due to the failures at level $0$, none of the state-of-the art solutions can generate useful subpolicies at higher levels. However, useful subpolices can be learned by DEHRL at higher levels. Fig.~\ref{sub-polices-level-1} visualizes five subpolicies learned by DEHRL at level~1, where four of them (marked by a green box) result in getting four different ingredients, which are the useful subpolicies at level $1$. 

\subsubsection{With Extrinsic Rewards.}




Although DEHRL can work in a bottom-up fashion where no extrinsic reward is required to discover useful subpolicies, DEHRL actually also has a very superior performance in the scenarios when extrinsic rewards are given. Therefore, we compare DEHRL with two state-of-the-art top-down methods,  \emph{option-critic} \cite{bacon2017option} and \emph{FeUdal} \cite{vezhnevets2017feudal}, where extrinsic rewards are essential. As shown in Table \ref{game-settings}, six different extrinsic reward settings are given to OverCooked, resulting in different difficulties.

To measure the performance quantitatively, two metrics, \emph{final performance score} and \emph{learning speed score}, which are based on reward per episode, are imported from \cite{schulman2017proximal}. Generally, the higher the reward per episode, the better the solution. Specifically, the final performance score averages the reward per episode over the last 100 episodes of training to measure the performance at final stages; while the learning speed score averages the extrinsic reward per episode over the entire training period to quantify the learning efficiency.

The results of the final performance score and the learning speed score are shown in Table \ref{final-performance-all} and Fig.~\ref{learning-speed-all}, respectively. In Table \ref{final-performance-all}, we find that DEHRL can solve the problems in all six settings, while option-critic can only solve the two easier ones. The failure of option-critic is because the extrinsic reward gets more sparse in the last four harder cases.
Besides,  Table \ref{final-performance-all}  shows that FeUdal fails when it is extended to 3 levels: its key idea “transition policy gradient” does not work well for multi-level structures, so it is hard to converge for $\textit{reward-levels}>2$.
Consequently, we state that DEHRL can also achieve a better performance than the state-of-the-art baselines when extrinsic rewards are given.

\begin{table}
	\centering
	\caption{\textrm{Final performance score of DEHRL 
			and baselines on OverCooked with six different extrinsic reward settings.}}
	
\smallskip 	
	\resizebox{\columnwidth}{!}{%
		\begin{tabular}{ccccccc}
			\midrule
			\begin{tabular}{@{}c@{}}\textit{reward-level} \\ \textit{goal-type}\end{tabular}  & \begin{tabular}{@{}c@{}}\textit{1} \\ \textit{any}\end{tabular} & \begin{tabular}{@{}c@{}}\textit{1} \\ \textit{fix}\end{tabular} & \begin{tabular}{@{}c@{}}\textit{1} \\ \textit{random}\end{tabular} & \begin{tabular}{@{}c@{}}\textit{2} \\ \textit{any}\end{tabular} & \begin{tabular}{@{}c@{}}\textit{2} \\ \textit{fix}\end{tabular} & \begin{tabular}{@{}c@{}}\textit{2} \\ \textit{random}\end{tabular} \\ \midrule
			DEHRL & \textbf{1.00}  & \textbf{1.00} & \textbf{1.00} & \textbf{0.95}  & \textbf{0.93} & \textbf{0.81} \\\hline
			Option-critic  & \textbf{1.00}  & \textbf{1.00} & 0.00 & 0.00  & 0.00 & 0.00 \\
			FeUdal  & \textbf{1.00}  & \textbf{1.00} & 0.93 & 0.00  & 0.00 & 0.00 \\\hline
			PPO & 0.98  & 0.97 & 0.56 & 0.00  & 0.00 & 0.00 \\
			State Novelty & \textbf{1.00}  & 0.96 & 0.95 & 0.00  & 0.00 & 0.00 \\
			Transition Novelty & \textbf{1.00}  & \textbf{1.00} & \textbf{1.00} & 0.00  & 0.00 & 0.00 \\ \hline
		\end{tabular}
	}
	\label{final-performance-all}
\end{table}

\begin{figure}
	\centering
		\centerline{\includegraphics[page=1,width=1\columnwidth]{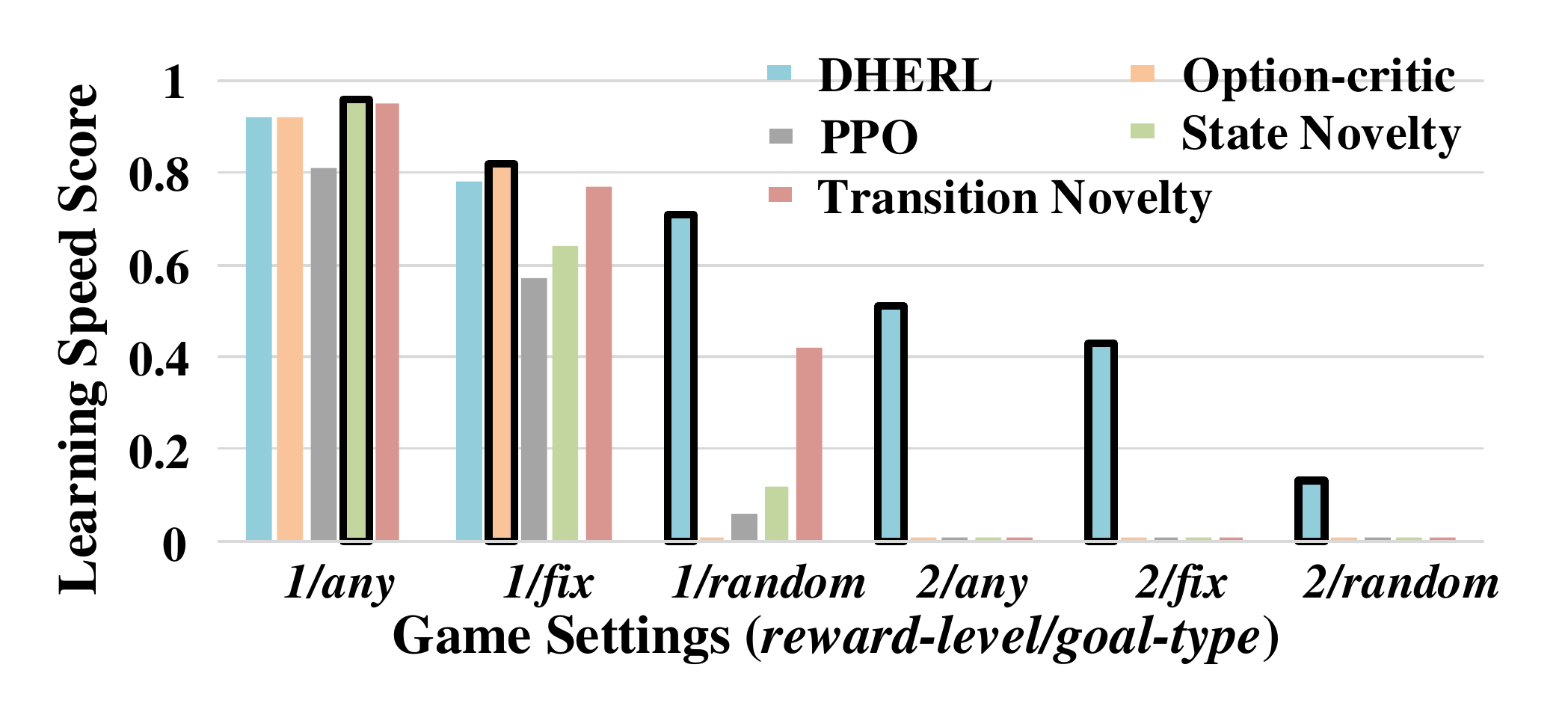}}
		\caption{\textrm{Learning speed score of DEHRL 
				and the baselines on OverCooked with six different extrinsic reward  settings.}}
		\label{learning-speed-all}
\end{figure}


\subsection{Solving the Sparse Extrinsic Reward Problem}

As option-critic fails in the sparse extrinsic reward problem, to illustrate the advantage of our method in solving this problem, we combine DEHRL with two state-of-the-art methods with better exploration strategies, namely, \emph{state novelty} \cite{csimcsek2004using} and \emph{transition novelty} \cite{pathak2017curiosity}. In addition, as previously mentioned, our framework is based on the \emph{PPO} algorithm \cite{schulman2017proximal}, so we include PPO as a baseline as well. The evaluations are also based on the final performance scores and learning speed scores, which are shown in Table \ref{final-performance-all} and Fig.~\ref{learning-speed-all}, respectively.
The results show that, better than option-critic, the three new baselines are all able to solve the task in the third setting. However, they all still fail in the last three settings. 

\subsection{Meta HRL}

HRL has recently shown a promising ability to learn meta-subpo\-li\-cies that better facilitate an adaptive behavior for new problems \cite{solway2014optimal,frans2017meta}.
We compare DEHRL against the state-of-the-art MLSH framework in \cite{frans2017meta} to investigate such a performance.

\begin{figure}
	\centering
	\subfigure{
		\includegraphics[page=2,width=\columnwidth]{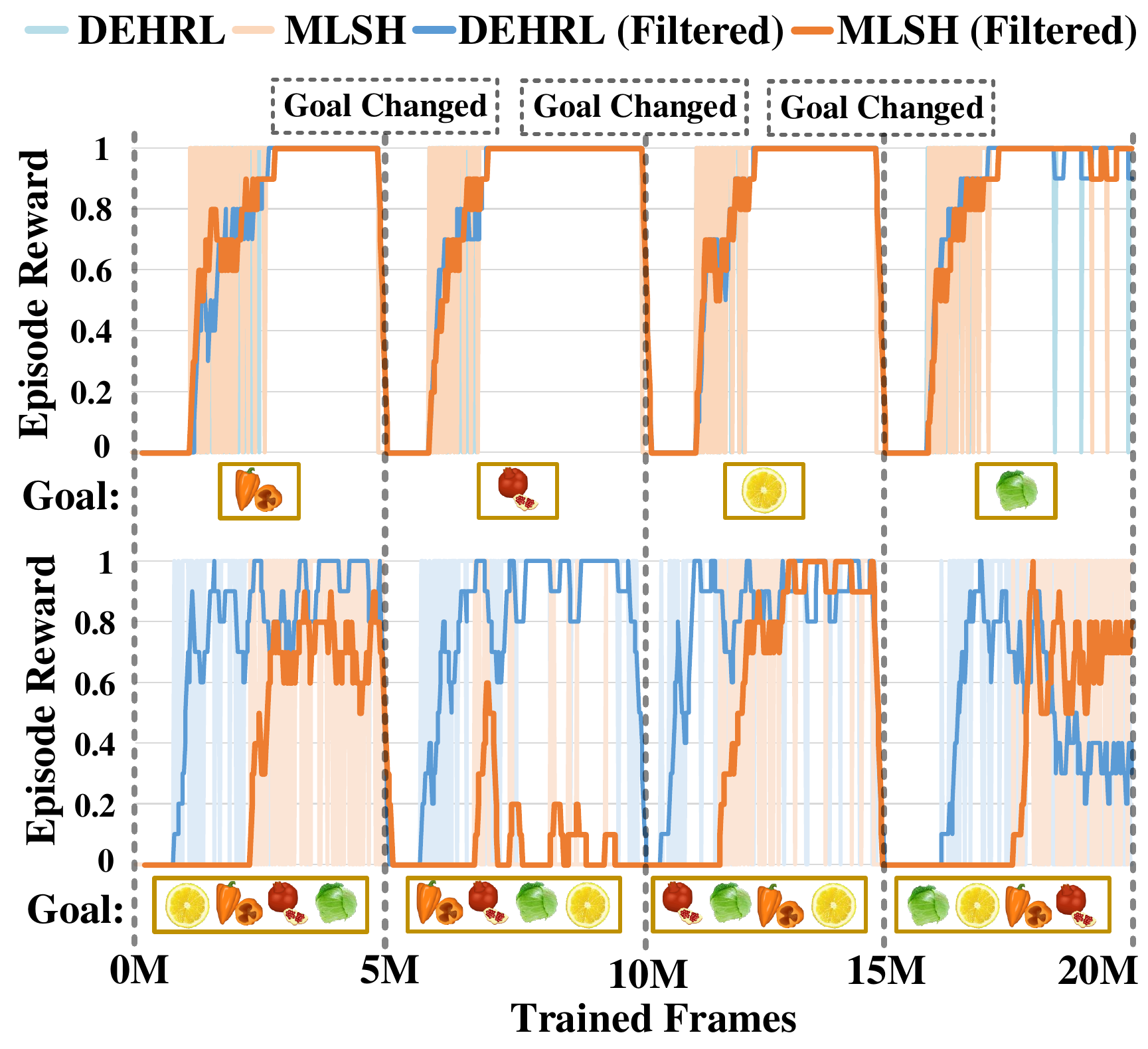}
	}
	\caption{\textrm{Meta HRL performances of DEHRL and MLSH in OverCooked with reward-level=1 (upper) and reward-level=2 (downer)}.}
	\label{meta-curve-level}
\end{figure}

We first test both DEHRL and MLSH in OverCooked with \textit{reward-level=1} and \textit{goal-type=random}.
In order to make MLSH work properly, instead of changing the goal every episode, as originally designed in \textit{goal-type=random}, the goal in MLSH is changed every 5M steps; for a fair comparison, the top-level hierarchy of DEHRL is also reset every 5M steps (same as MLSH). The results of the episode extrinsic reward curves in 20M steps (the goal is changed five times) are shown in Fig.~\ref{meta-curve-level} (upper part).
As expected, the episode extrinsic reward drops every time the goal is changed, since the top-level hierarchies of both methods are re-initialized. The increase speed of the episode extrinsic reward after each reset can measure the performance of methods in learning meta-subpolicies that facilitate an adaptive behavior for a new goal.
Consequently, we find that DEHRL and MLSH have a similar meta-learning performance under this setting.

In addition, we repeat the above experiment with \textit{reward-level=2}, and the results are shown in Fig.~\ref{meta-curve-level} (lower part). We find that DEHRL produces a better meta-learning ability than MLSH. This is because DEHRL is capable to learn the subpolicies at level $1$ to fetching four different ingredients, while MLSH can only learn subpolicies to moving towards four different directions (similar to the subpolicies at level $0$ of DEHRL). Obviously, based on the better intertask-transferable subpolicies learned at level $1$, DEHRL will resolve the new goal more easily and quickly than MLSH. Thus, this finding shows that DEHRL can learn meta subpolicies at higher levels, which are usually better intertask-transferable than those learned by the baseline.


\begin{figure}
	\centering
		\centerline{\includegraphics[page=1,width=1.1\columnwidth]{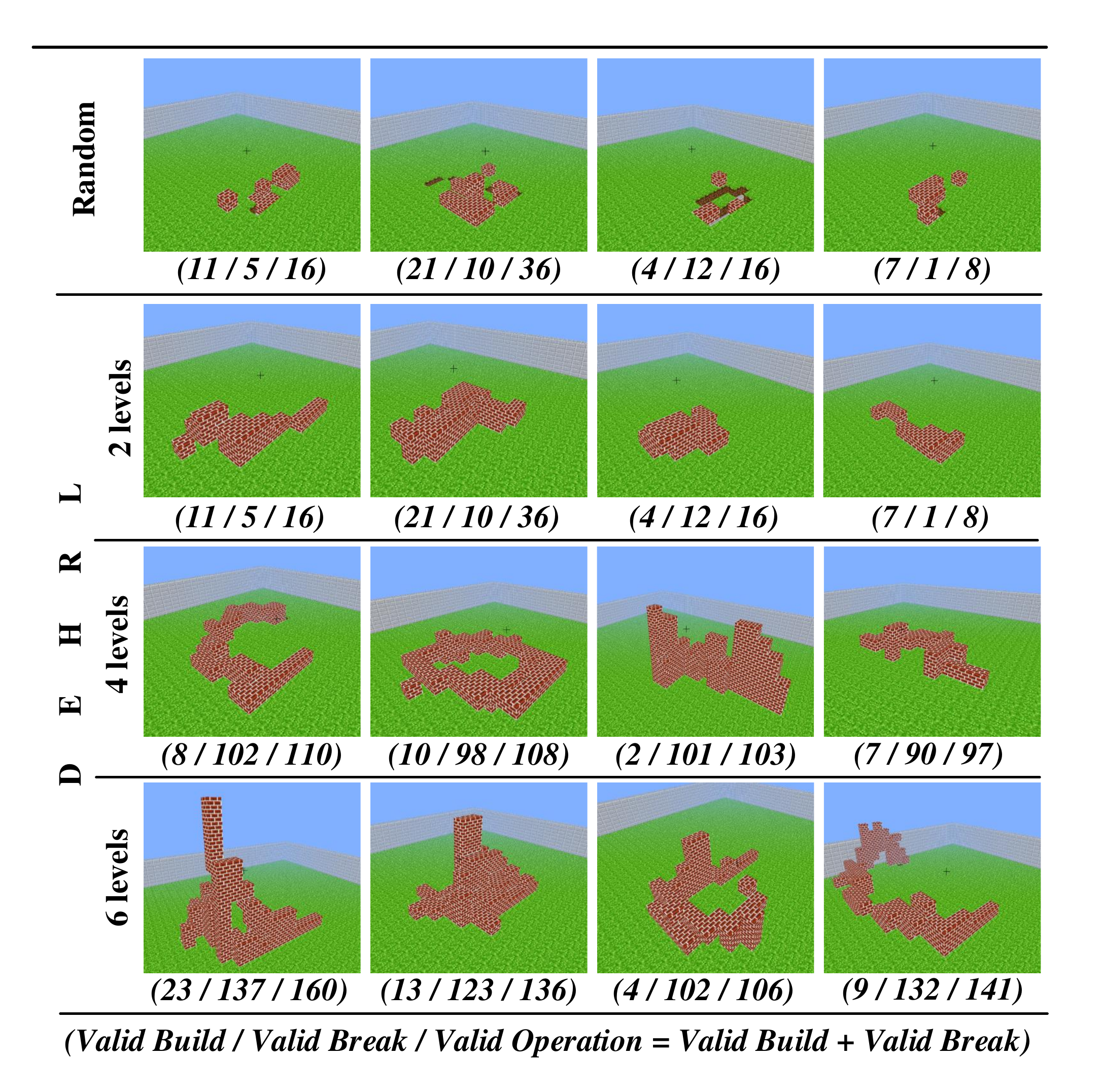}}
		\caption{\textrm{Worlds built by playing MineCraft without extrinstic reward.}}
		\label{minecraft}
\end{figure}


\subsection{Application of DEHRL in MineCraft}

To show the portability of DEHRL, we further apply DEHRL in a popular video game called \emph{MineCraft}, where the agent has much freedom to act and build.

In our settings, the agent has the first-person view via raw pixel input.
At the beginning of each episode, the world is empty except one layer of \textit{GRASS} blocks that can be broken.
We allow the agent to play 1000 steps in each episode; then the world is reset. At each step, ten actions are available, i.e., moving towards four directions, rotating the view towards four directions, break and build a block\footnote{Only one kind of block (i.e., \emph{BRICK}) can be built, and any blocks except \emph{STONE} can be broken.}, and jump. Due to space limits, more detailed settings and more visualizations about the agent and this game are provided in the arXiv release.


Since the typical use of DEHRL is based on the intrinsic reward only, the existing work \cite{tessler2017deep} that requires human-designed extrinsic reward signals to train subpolicies is not applicable to be used as baseline. Consequently, we compare the performance of DEHRL with three different numbers of levels in MineCraft to a framework with a random policy. Fig.~\ref{minecraft} shows the building results, where we measure the performance by world complexity. As we can see, DEHRL builds more complex worlds than the random policy. Furthermore, with the increase of the number of levels, DEHRL tends to build more complex worlds.

The complexity of the worlds is quantified by \emph{Valid Operation}, which is computed by the following equation:
\begin{equation*}
Valid~Operations = Valid~Breaks +  Valid~Builds,
\end{equation*}
\noindent where \emph{Valid Builds} is the number of blocks that have been built and not broken at the end of an episode; \emph{Valid Breaks} is the number of blocks that are originally in the world that have been broken.
Consequently, blocks that are built but broken later will not be counted into \textit{Valid Build} or \textit{Valid Break}.
The quantitative results in Fig.~\ref{minecraft} are definitely consistent with our previous intuitive feeling.

\begin{figure*}
	\centering
	\includegraphics[page=1,width=0.8\textwidth]{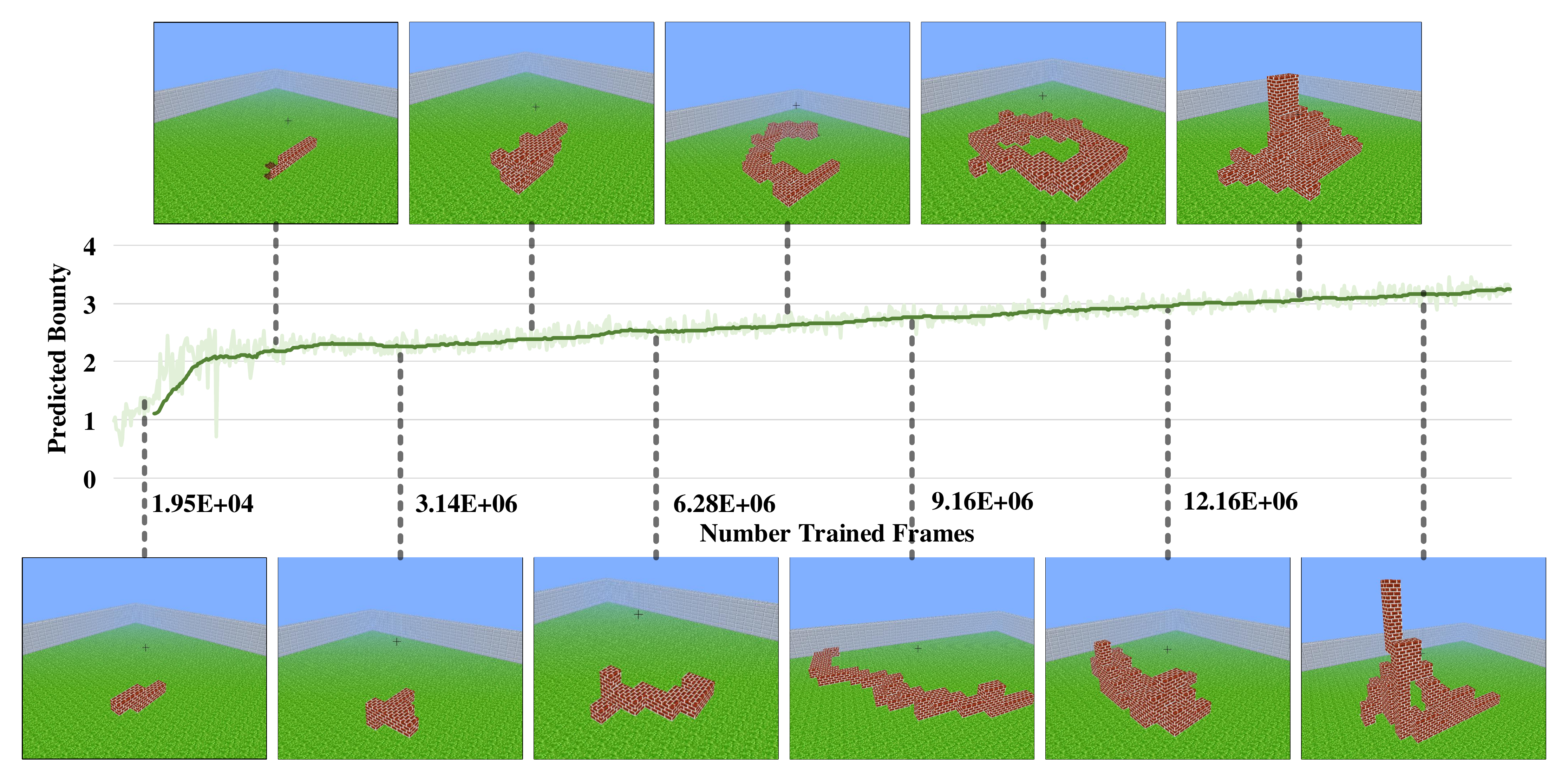}
	\caption{Predicted intrinsic reward.}
	\label{intrinsic-reward-clip}
\end{figure*}

Finally, since the predicted intrinsic reward ($\hat{b}_t^l$) is an indication of the diversity of current subpolicies, we plot the averaged $\hat{b}_t^l$ over all levels and visualize the world built by DEHRL at the point in Fig.~\ref{intrinsic-reward-clip}, so that the relationship between $\hat{b}_t^l$ and the built world is further illustrated.

\section{Related Work}

\smallskip
\noindent\textbf{Discovering diverse subpolicies in HRL.}
The diversity assumption is prevailing in the recent works of option discovery.
Among them, SAC-LSP \cite{haarnoja2018latent} is the most recent work, but whether it can operate different layers at different temporal scales is an open problem.
HiREPS \cite{daniel2012hierarchical} is also a popular approach, but it works in a top-down fashion.
Thus, it is not clear  whether the above two methods can be applied to sparse extrinsic reward tasks.
In contrast, SNN \cite{florensa2017stochastic} is designed to handle sparse extrinsic reward tasks, which achieves this diversity assumption explicitly by information-maximizing statistics.
Besides, it is promising to apply SNN in HRL of multiple levels.
However, SNN suffers from various failures when the possible future states are enormous, making it impractical on domains with a large action space and unable to further learn higher-level subpolicies.
Similar failure cases are observed in \cite{gregor2016variational}.

\smallskip
\noindent\textbf{Extensible HRL. }%
Recently, there has been some attempts in increasing the levels of HRL.
Such works include MAXQ \cite{dietterich2000hierarchical}, which requires completely searching the subtree of each subtask, leading to high computational costs.
In contrast, AMDP \cite{gopalan2017planning} explores only the relevant branches.
However, AMDP concentrates on the planning problem.
Deeper levels are also supported in \cite{silver2012compositional}, but its scalability is not clear.
DDO \cite{fox2017multi} and DDCO \cite{krishnan2017ddco} discover higher-level subpolicies from demonstration trajectories.
However, our work focuses on learning those purely end-to-end without human-designed extrinsic reward.
Other works \cite{rasmussen2017neural,song2017generalization} also involve a modular structure that supports deeper-level HRL.
However, there is no guarantee or verification on whether the structure can learn useful or diverse subpolicies at different temporal scales.

\smallskip
\noindent\textbf{Meta HRL. }%
Neuroscience research \cite{solway2014optimal} proposes that the optimal hierarchy is the one that best facilitates an adaptive behavior in the face of new problems.
Its idea is accomplished with a verified scalability in MLSH \cite{frans2017meta}, where meta HRL is proposed.
However, MLSH keeps reinitializing the policy at the top level, once the environment resets the goal. This brings several drawbacks, such as requiring auxiliary  information from the environment about when the goal has been changed. In contrast, our method does not introduce such a restriction.
Regardless of the difference, we compare with MLSH in our experiments under the settings of MLSH, where auxiliary  information on goal resetting is provided. As such, the meta HRL ability of our approach is investigated.

\smallskip
\noindent\textbf{Improved exploration with predictive models. }%
Since we introduce the transition model to generate intrinsic rewards, our method is also related to RL improvements with predictive models, typically introducing sample models \cite{fu2017ex2}, generative models \cite{song2017learning}, or deterministic models \cite{pathak2017curiosity} as transition models to predict future states.
However, the transition model in our DEHRL is introduced to encourage developing diverse subpolicies, while those in the above works are introduced to improve the exploration.
Our method is compared with the above state novelty \cite{csimcsek2004using} and transition novelty \cite{pathak2017curiosity} in our experiments.

\section{Summary and Outlook}

We have proposed DEHRL towards building extensible HRL that learns useful subpolicies over multiple levels efficiently.
However, there are several interesting directions to explore.
One of them is to develop algorithms that generate or dynamically adjust the settings of $T^l$ and $\mathbb{A}^l$.
Furthermore, measuring the distance between states is another important direction to explore, where better representations of states may lead to improvements.
Finally, DEHRL may be a promising solution for visual tasks \cite{xu2018predicting} with diverse representation and mixed reward functions.

\subsubsection{Acknowledgments.}

This work was supported by the State Scholarship Fund awarded by China Scholarship Council and 
by the Alan Turing Institute under the UK EPSRC grant EP/N510129/1. 

\section{Supplementary Material}

\begin{table}[H]
	\centering
	\begin{tabular}{c|c}
		\midrule
		\textit{Hyperparameters} & \textit{Value}\\
		\midrule
		Horizon (T) & 128 \\
		Adam stepsize & $2.5\times10^{-4}\times2$ \\
		Learning rate & $7\times10^4$\\
		Number epochs & 4\\
		Minibatch size & $32\times8$ \\
		Discount ($\gamma$) & 0.99\\
		GAE parameter ($\lambda$) & 0.95 \\
		Number of actors & 8 \\
		Clipping parameter ($\epsilon$) & $0.1 \times 2$ \\
		VF coefficient ($c^1$) & 0.5\\
		Entropy coefficient ($c^2$) & 0.01 \\
		\midrule
	\end{tabular}
	\caption{PPO hyperparameters used for DEHRL at each level on OverCooked.}
	\label{hyper_table_OverCooked}
\end{table}

\begin{table}[H]
	\centering
	\begin{tabular}{c|c}
		\midrule
		\textit{Hyperparameters} & \textit{Value}\\
		\midrule
		Horizon (T) & 128 \\
		Adam stepsize & $2.5\times10^{-4}\times2$ \\
		Learning rate & $7\times10^4$\\
		Number epochs & 4\\
		Minibatch size & $32\times8$ \\
		Discount ($\gamma$) & 0.99\\
		GAE parameter ($\lambda$) & 0.95 \\
		Number of actors & 1 \\
		Clipping parameter ($\epsilon$) & $0.1 \times 2$ \\
		VF coefficient ($c^1$) & 0.5\\
		Entropy coefficient ($c^2$) & 0.01 \\
		\midrule
	\end{tabular}
	\caption{PPO hyperparameters used for DEHRL at each level on MineCraft.}
	\label{hyper_table_MineCraft}
\end{table}

\begin{table}\renewcommand{\arraystretch}{1.1}
	\centering
	\begin{tabular}{cccc}
		\midrule
		$l$            & 0  & 1     & 2 \\
		$\mathbb{A}^l$ & 16 & 5     & 5  \\
		$T^l$          & 1  & $1*4$ & $1*4*12$ \\ \hline
	\end{tabular}%
	\caption{DEHRL Settings on OverCooked.}
	\label{dehrl-settings-overcooked}
\end{table}

\begin{table}\renewcommand{\arraystretch}{1.1}
	\centering
	\begin{tabular}{ccccccc}
		\midrule
		$l$            & 0  & 1     & 2       & 3         & 4           & 5 \\
		$\mathbb{A}^l$ & 11 & 8     & 8       & 8         & 8           & 8 \\
		$T^l$          & 1  & $1*4$ & $1*4^2$ & $1*4^3$ & $1*4^4$ & $1*4^5$   \\ \hline
	\end{tabular}%
	\caption{DEHRL Settings on MineCraft.}
	\label{dehrl-settings-minecraft}
\end{table}

\begin{table*}
	\centering
	\begin{tabular}{c|c}
		\midrule
		\textbf{Input 1}: current \textit{state} ($s_t$), as $84\times84$ gray scaled image &  \textbf{Input 2}: \textit{action} from level $l+1$ ($a_t^{l+1}$), as one-hot vector \\
		\midrule
		\multicolumn{2}{c}{\textbf{Conv}: kernel size $8\times8$, number of features 16, stride 4} \\
		\multicolumn{2}{c}{\textbf{lRELU}} \\
		\multicolumn{2}{c}{\textbf{Conv}: kernel size $4\times4$, number of features 32, stride 2} \\
		\multicolumn{2}{c}{\textbf{lRELU}} \\
		\multicolumn{2}{c}{\textbf{Conv}: kernel size $3\times3$, number of features 16, stride 1} \\
		\multicolumn{2}{c}{\textbf{lRELU}} \\
		\multicolumn{2}{c}{\textbf{Flatten}: $16\times16\times7$ is flatten to $1792$} \\
		\multicolumn{2}{c}{\textbf{FC}: number of features 256}\\
		\multicolumn{2}{c}{\textbf{lRELU}} \\
		\midrule
		\textbf{Output 1}: multiple policy functions, each one for one $a_t^{l+1}$ &  \textbf{Output 2}: multiple value functions, each one for one $a_t^{l+1}$ \\
		\midrule
	\end{tabular}
	\caption{Network architecture of the \textit{policy} at each level ($\pi^l$).}
	\label{network_policy}
\end{table*}

\begin{table*}
	\centering
	\begin{tabular}{c|c}
		\midrule
		\textbf{Input 1}: current \textit{state} ($s_t$), as $84\times84$ gray scaled image &  \textbf{Input 2}: \textit{action} from level $l$ ($a_t^{l}$), as one-hot vector \\
		\midrule
		\textbf{Conv}: kernel size $8\times8$, number of features 16, stride 4 & \multirow{11}*{\begin{tabular}{@{}c@{}}\textbf{FC}: number of features 256 \end{tabular} } \\
		\textbf{BatchNorm} & \\
		\textbf{lRELU} \\
		\textbf{Conv}: kernel size $4\times4$, number of features 32, stride 2 & \\
		\textbf{BatchNorm} & \\
		\textbf{lRELU} \\
		\textbf{Conv}: kernel size $3\times3$, number of features 16, stride 1 & \\
		\textbf{BatchNorm} & \\
		\textbf{lRELU} \\
		\textbf{Flatten}: $16\times16\times7$ is flatten to $1792$ & \\
		\textbf{FC}: number of features 256 & \\
		\midrule
		\multicolumn{2}{c}{\textbf{Dot-multiply}}\\
		\midrule
		\textbf{FC}: number of features 256 & \multirow{11}*{\begin{tabular}{@{}c@{}}\textbf{FC}: number of features 1 \end{tabular}  }\\
		\textbf{FC}: number of features $1792$ & \\
		\textbf{Reshape}: $1792$ is reshaped to $16\times16\times7$  & \\
		\textbf{DeConv}: kernel size $3\times3$, number of features 32, stride 1 &\\
		\textbf{BatchNorm} & \\
		\textbf{lRELU} \\
		\textbf{DeConv}: kernel size $4\times4$, number of features 16, stride 2 &\\
		\textbf{BatchNorm} & \\
		\textbf{lRELU} \\
		\textbf{DeConv}: kernel size $8\times8$, number of features 1, stride 4 &\\
		\textbf{Sigmoid} \\
		\midrule
		\textbf{Output 1}: predicted \textit{state} after $T^l$ steps ($\hat{s}_{t+T^l}$) &  \textbf{Output 2}: predicted \textit{bounty} at downer level ($\hat{b}^{l-1}_t$) \\
		\midrule
	\end{tabular}
	\caption{Network architecture of the \textit{predictor} at each level ($\beta^l$).}
	\label{network_predictor}
\end{table*}

\begin{table*}\renewcommand{\arraystretch}{1.1}
	\centering
	\begin{tabular}{ccccccc}
		\midrule
		\textit{reward-level} / \textit{goal-type}  & \textit{1} / \textit{any} & \textit{1} / \textit{fix} & \textit{1} / \textit{random} & \textit{2} / \textit{any} & \textit{2} / \textit{fix} & \textit{2} / \textit{random} \\
		\midrule
		DHERL & \textbf{1.00}  & \textbf{1.00} & \textbf{1.00} & \textbf{0.95}  & \textbf{0.93} & \textbf{0.81} \\
		Option-critic\cite{bacon2017option}  & \textbf{1.00}  & \textbf{1.00} & 0.00 & 0.00  & 0.00 & 0.00 \\
		PPO\cite{schulman2017proximal} & 0.98  & 0.97 & 0.56 & 0.00  & 0.00 & 0.00 \\
		State Novelty\cite{csimcsek2004using} & \textbf{1.00}  & 0.96 & 0.95 & 0.00  & 0.00 & 0.00 \\
		Transition Novelty\cite{pathak2017curiosity} & \textbf{1.00}  & \textbf{1.00} & \textbf{1.00} & 0.00  & 0.00 & 0.00 \\ \hline
	\end{tabular}
	\caption{\textrm{Final performance of DHERL, Option-critic\cite{bacon2017option}, PPO\cite{schulman2017proximal}, State Novelty\cite{csimcsek2004using} and Transition Novelty\cite{pathak2017curiosity} on OverCooked of 6 settings.}}
	\label{final-performance-all}
\end{table*}

\begin{table*}\renewcommand{\arraystretch}{1.1}
	\centering
	\begin{tabular}{ccccccc}
		\midrule
		\textit{reward-level} / \textit{goal-type}  & \textit{1} / \textit{any} & \textit{1} / \textit{fix} & \textit{1} / \textit{random} & \textit{2} / \textit{any} & \textit{2} / \textit{fix} & \textit{2} / \textit{random} \\
		\midrule
		DHERL & 0.92 & 0.72 & \textbf{0.71} & \textbf{0.51}  & \textbf{0.43} & \textbf{0.13} \\
		Option-critic\cite{bacon2017option}  & 0.92  & \textbf{0.82} & 0.00 & 0.00  & 0.00 & 0.00 \\
		PPO\cite{schulman2017proximal} & 0.81  & 0.57 & 0.06 & 0.00  & 0.00 & 0.00 \\
		State Novelty\cite{csimcsek2004using} & \textbf{0.96}  & 0.64 & 0.12 & 0.00  & 0.00 & 0.00 \\
		Transition Novelty\cite{pathak2017curiosity} & 0.95  & 0.77 & 0.42 & 0.00  & 0.00 & 0.00 \\ \hline
	\end{tabular}
	\caption{\textrm{Learning Speed of DHERL, Option-critic\cite{bacon2017option}, PPO\cite{schulman2017proximal}, State Novelty\cite{csimcsek2004using} and Transition Novelty\cite{pathak2017curiosity} on OverCooked of 6 settings.}}
	\label{learning-speed-all}
\end{table*}

\begin{figure*}
	\centering
	\begin{tabular}{cc}
		\includegraphics[page=1,width=0.4\textwidth]{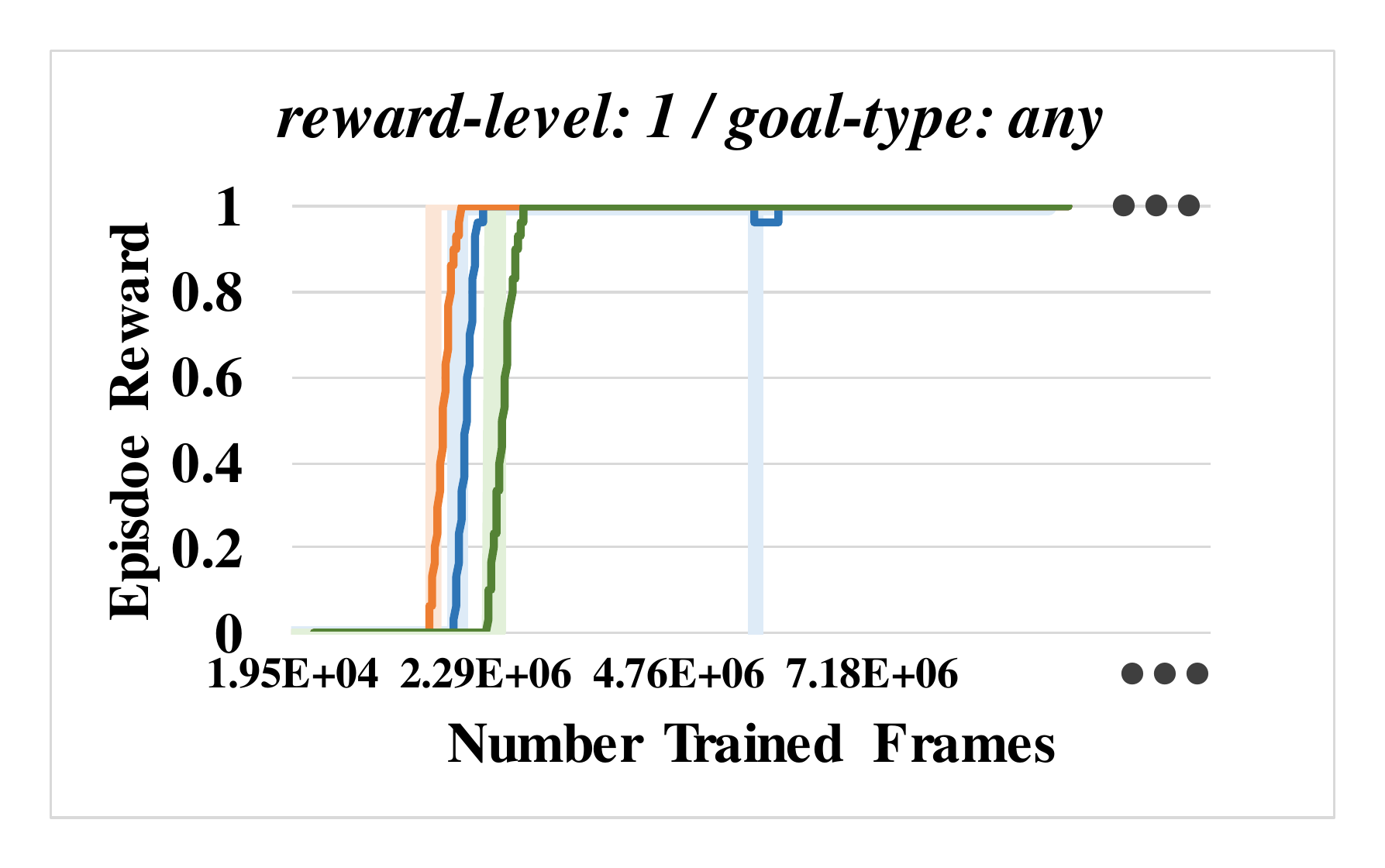}
		&
		\includegraphics[page=4,width=0.4\textwidth]{episode-reward-curve.pdf}
		\\
		\includegraphics[page=2,width=0.4\textwidth]{episode-reward-curve.pdf}
		&
		\includegraphics[page=5,width=0.4\textwidth]{episode-reward-curve.pdf}
		\\
		\includegraphics[page=3,width=0.4\textwidth]{episode-reward-curve.pdf}
		&
		\includegraphics[page=6,width=0.4\textwidth]{episode-reward-curve.pdf}
		\\
	\end{tabular}
	\caption{Episode reward curve of DEHRL on all 6 settings of OverCooked. Different colors indicate runs with different training seeds. Shallower color indicates the original curve and the darker color indicates the filtered curve.}
	\label{episode-reward-curve}
\end{figure*}

\begin{figure*}
	\centering
	\subfigure[Predicted states by \textit{predictor} at level 1]{
		\includegraphics[page=1,width=0.7\textwidth]{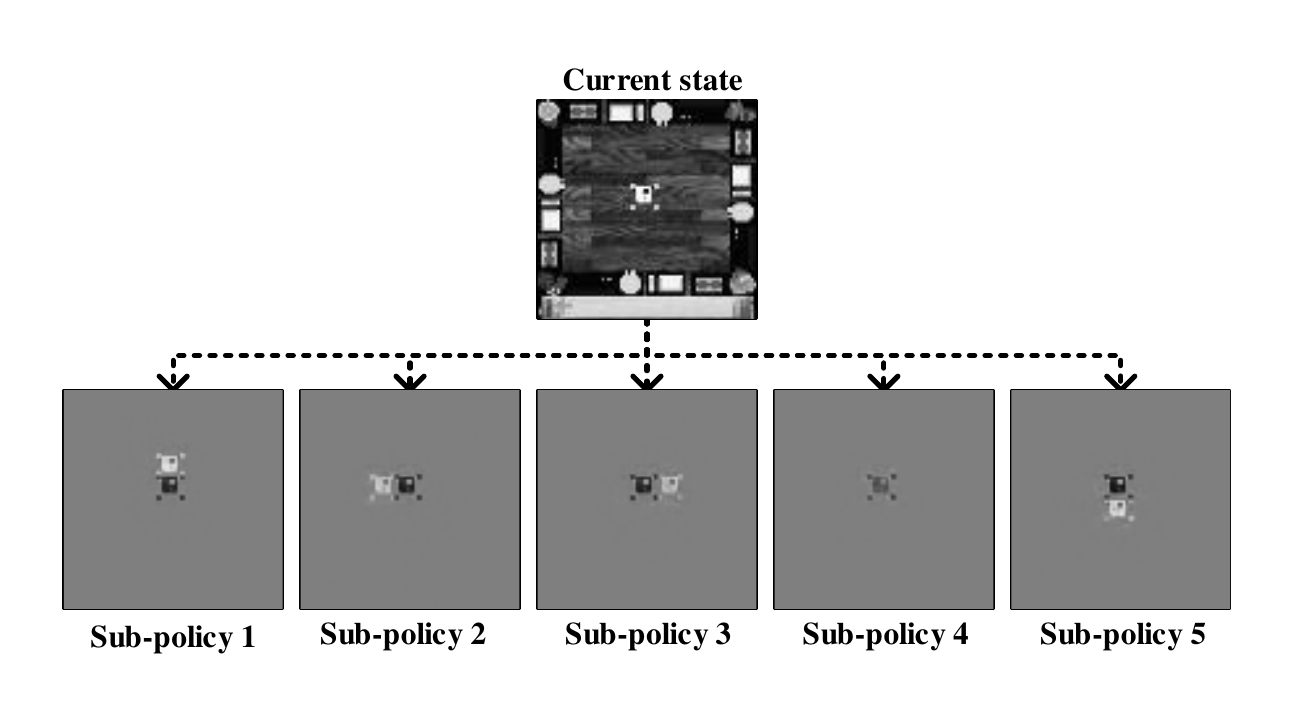}
	}
	\subfigure[Predicted states by \textit{predictor} at level 2]{
		\includegraphics[page=2,width=0.7\textwidth]{transition-prediction.pdf}
	}
	\caption{Predicted states by predictor at each level on OverCooked.}
	\label{transition-prediction}
\end{figure*}

\begin{figure*}
	\centering
	\begin{tabular}{cccc}
		\includegraphics[page=1,width=0.225\textwidth]{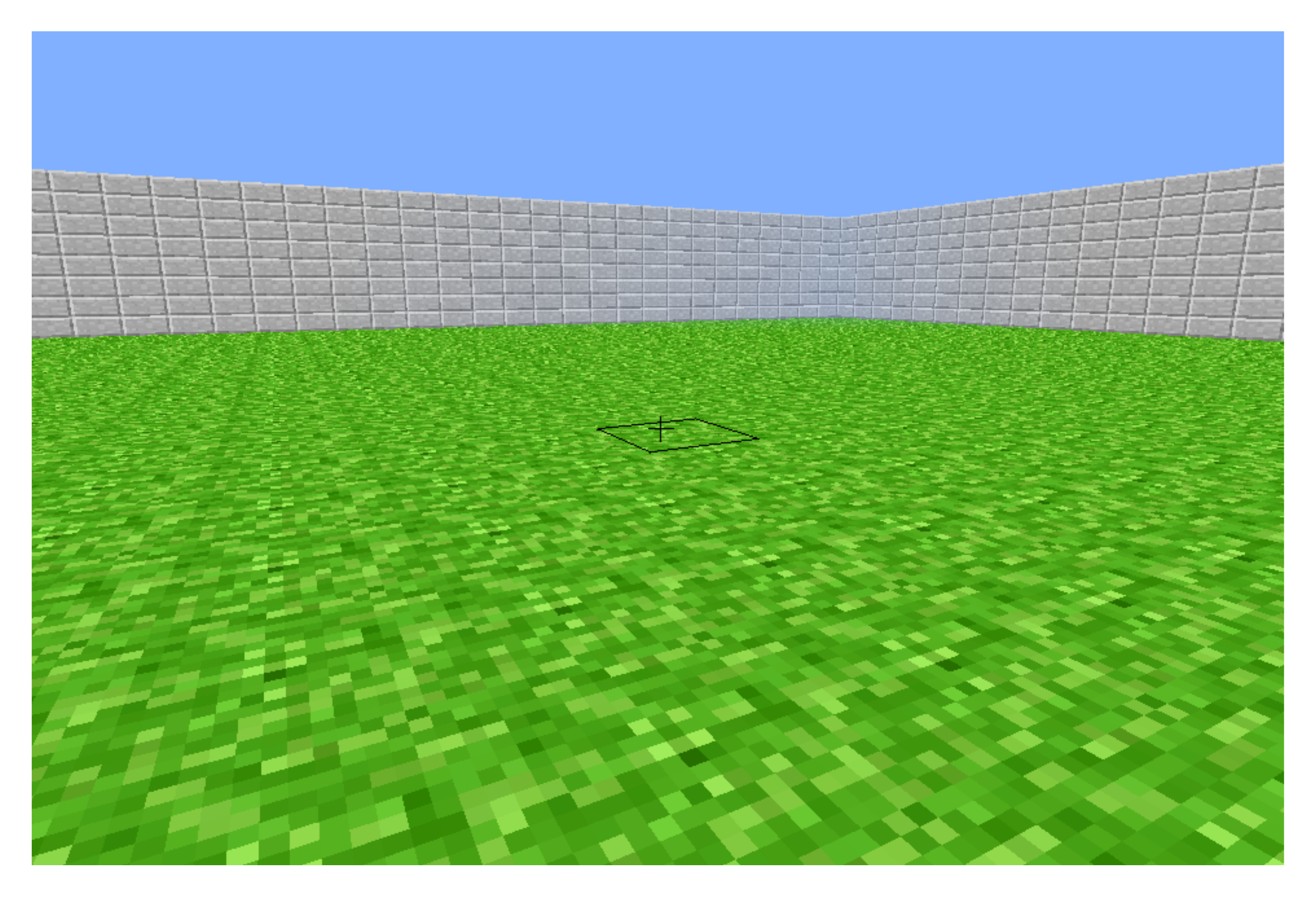}
		&
		\includegraphics[page=2,width=0.225\textwidth]{example-states-minecraft.pdf}
		&
		\includegraphics[page=3,width=0.225\textwidth]{example-states-minecraft.pdf}
		&
		\includegraphics[page=4,width=0.225\textwidth]{example-states-minecraft.pdf}
		\\
		(a) Start state & (b) Break a block & (c) Build a block & (d) Jump on a block \\
	\end{tabular}
	\caption{Example states in MineCraft.}
	\label{example-states-minecraft}
\end{figure*}

\subsection{Hyperparameters}
The \textit{policy} at each level is trained with Proximal Policy Optimization (PPO) algorithm \cite{schulman2017proximal}.
Detailed settings of the hyper-parameters are shown in Table \ref{hyper_table_OverCooked} and \ref{hyper_table_MineCraft} for OverCooked and MineCraft respectively.
Detailed settings of DEHRL framework for OverCooked and MineCraft are shown in Table \ref{dehrl-settings-overcooked} and \ref{dehrl-settings-minecraft} respectively.

\subsection{Neural Network Details}

The details of network architecture for \textit{policy} and \textit{predictor} at each level is shown in Table \ref{network_policy} and \ref{network_predictor} respectively.
Fully connected layer is denoted as FC and flatten layer is denoted as Flatten.
We use leaky rectified linear units (denoted as LeakyRELU) \cite{maas2013rectifier} with
leaky rate $0.01$ as the nonlinearity applied to all the hidden layers in our network.
Batch normalization \cite{ioffe2015batch} (denoted as BatchNorm) is applied after hidden convolutional layers (denoted as Conv) in \textit{predictor}.
For the \textit{predictor} at each level, the integration of the two inputs, i.e., \textit{state} and \textit{action}, is accomplished by approximated multiplicative interaction \cite{oh2015action} (the \textbf{dot-multiply} operation in Table \ref{network_predictor}), so that any predictions made by the \textit{predictor} are conditioned on the \textit{action} input.
Deconvolutional layers (denoted as DeConv) \cite{zeiler2011adaptive} are applied for predicting the state after $T^l$ steps.

\subsubsection{Performance on OverCooked}

Here we include a comparison of DEHRL against Option-critic\cite{bacon2017option}, PPO\cite{schulman2017proximal}, State Novelty\cite{csimcsek2004using} and Transition Novelty\cite{pathak2017curiosity} on OverCooked of 6 settings.
Table \ref{final-performance-all} shows the final performance and Table \ref{learning-speed-all} shows the learning speed.

Besides, there is an interesting question to answer for SNN \cite{florensa2017stochastic}: If SNN is guaranteed to learn different subpolicies, will it learn the 5 useful ones provided with enough subpolicy models (set $\mathbb{A}^1=625$)?
We train the above settings for 200M steps with 3 trials of different training seeds.
Surprisingly, the best trial learns 2 useful subpolicies.
The reason is that setting $\mathbb{A}^1=625$ makes the estimation of the mutual information easily inaccurate, since the mutual information is estimated for every $a^1_t$ in $\mathbb{A}^1$.

Figure \ref{episode-reward-curve} shows the learning curves of DEHRL with three different training seeds.

Figure \ref{transition-prediction} shows the predicted states of the \textit{predictor} at each level in DEHRL.
Since the size of observations and the predictions is $84\time84$ and they are gray scaled, it would be hard to have a clear visualization of the predictions.
Thus, just for better visualization, current observation is subtracted from the predictions to remove the unchanged parts in Figure \ref{transition-prediction}.

\subsubsection{Performance on MineCraft}

Figure \ref{example-states-minecraft} shows the example states observed by the \textit{agent} in MineCraft.



\bibliography{yuhangsong}

\bibliographystyle{aaai}
\end{document}